\documentclass[sigconf,nonacm]{acmart}
\usepackage{caption}

\AtBeginDocument{%
  }

\begin{document}

\title[Work-in-Progress]{TeachingCoach: A Fine-Tuned Scaffolding Chatbot for Instructional Guidance to Instructors}

\author{Isabel Molnar}
\authornote{Both authors contributed equally to this work.}
\email{imolnar@nd.edu}
\affiliation{%
  \institution{University of Notre Dame}
  \city{Notre Dame}
  \state{Indiana}
  \country{USA}
}

\author{Peiyu Li}
\authornotemark[1]
\email{pli9@nd.edu}
\affiliation{%
  \institution{University of Notre Dame}
  \city{Notre Dame}
  \state{Indiana}
  \country{USA}
}

\author{Si Chen}
\email{schen34@nd.edu}
\affiliation{%
  \institution{University of Notre Dame}
  \city{Notre Dame}
  \state{Indiana}
  \country{USA}
}

\author{Sugana Chawla}
\email{schawla@nd.edu}
\affiliation{%
  \institution{University of Notre Dame}
  \city{Notre Dame}
  \state{Indiana}
  \country{USA}
}

\author{James Lang}
\email{jlang4@nd.edu}
\affiliation{%
  \institution{University of Notre Dame}
  \city{Notre Dame}
  \state{Indiana}
  \country{USA}
}

\author{Ronald Metoyer}
\email{rmetoyer@nd.edu}
\affiliation{%
  \institution{University of Notre Dame}
  \city{Notre Dame}
  \state{Indiana}
  \country{USA}
}

\author{Ting Hua}
\email{thua@nd.edu}
\affiliation{%
  \institution{University of Notre Dame}
  \city{Notre Dame}
  \state{Indiana}
  \country{USA}
}

\author{Nitesh V. Chawla}
\email{nchawla@nd.edu}
\affiliation{%
  \institution{University of Notre Dame}
  \city{Notre Dame}
  \state{Indiana}
  \country{USA}
}




\renewcommand{\shortauthors}{Molnar*, Li* et al.}

\begin{abstract}

Higher education instructors often lack timely and pedagogically grounded support, as scalable instructional guidance remains limited and existing tools rely on generic chatbot advice or non-scalable teaching center human-human consultations. We present \emph{TeachingCoach}, a pedagogically grounded chatbot designed to support instructor professional development through real-time, conversational guidance. TeachingCoach is built on a data-centric pipeline that extracts pedagogical rules from educational resources and uses synthetic dialogue generation to fine-tune a specialized language model that guides instructors through problem identification, diagnosis, and strategy development. Expert evaluations show TeachingCoach produces clearer, more reflective, and more responsive guidance than a GPT-4o mini baseline, while a user study with higher education instructors highlights trade-offs between conversational depth and interaction efficiency. Together, these results demonstrate that pedagogically grounded, synthetic data driven chatbots can improve instructional support and offer a scalable design approach for future instructional chatbot systems.

\end{abstract}



\maketitle

\section{Introduction}\label{sec1}

Effective teaching requires not only subject-matter expertise but also skillful use of instructional strategies. Although educational research provides abundant rules and principles, instructors often struggle to apply them systematically in classroom contexts. Universities operate teaching and learning centers that offer workshops and consultations, but these resources face key limitations: (1) support may not be available at the moment of need, (2) feedback is often generic rather than tailored to an instructor’s background, and (3) some educators hesitate to seek direct help for fear of appearing unskilled. As a result, many lack scalable and accessible support for their professional growth.  

Prior research on intelligent tutoring systems \cite{graesser2004autotutor,johnson2018pedagogical} shows that dialogue can scaffold reasoning and reflection, approaching the effectiveness of human tutors. More recently, educational chatbots and LLM-based frameworks  \cite{xu2024eduagent,shi2025educationq,liu2025pace,gao2025agent4edu} have been developed to model or support students, but surveys note that these systems are often weakly grounded in pedagogy and rarely address instructor professional development \cite{okonkwo2021chatbots,debets2025chatbots}.  

Recent work has focused on supporting instructors through AI-enabled analytics, assessment scalability, and workload reduction. For example, studies examined the use of AI-generated practice questions to scale established learning effects \cite{van2025scaling}, conversational interfaces that help instructors interpret learning analytics \cite{yang2025chat}, and LLM-based assignment report summaries that support teacher insight \cite{lim2025leveraging}. However, little work directly targets instructional support grounded in pedagogical practices for higher education instructors; instead, such support often relies on teaching center consultants, which is difficult to scale and frequently unavailable at smaller or less-resourced institutions, highly US-centric, where instructional support may be limited or entirely absent.\footnote{\url{https://podnetwork.org/content/uploads/Wright_PNN_NoCTLs_Jan2019_update2pdf.pdf}}

\begin{figure*}[t]
\centering
\includegraphics[width=\linewidth,trim=0 0 0 0,clip]{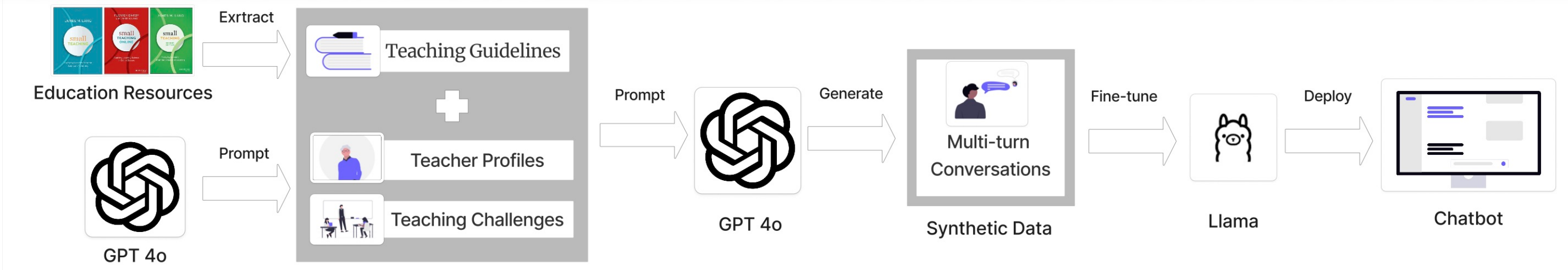}
\caption{Pipeline of TeachingCoach. Teaching guidelines are extracted from education resources, while GPT-4o generates teacher profiles and teaching challenges. These inputs prompt GPT-4o to produce synthetic multi-turn conversations, which, after expert filtering, are used to fine-tune a LLaMA model deployed as the TeachingCoach chatbot.}
\vspace{-1em}
\label{fig:pipeline}
\end{figure*}

Meanwhile, general-purpose LLMs like ChatGPT have become widely accessible, but their responses are typically generic and rarely apply evidence-based principles in a pedagogically scaffolded way. To bridge this gap, we introduce \textit{TeachingCoach}, a chatbot system that simulates the role of a teaching expert and delivers conversational guidance on instructional practice. At its core is a pipeline that grounds the system in evidence-based pedagogy while enabling scalable data generation and model training. Figure~\ref{fig:pipeline} illustrates this process, which we detail in the following sections.

\section{Related Works}

Large language models (LLMs) have been widely used as \emph{data generators} to synthesize labeled examples and structured interactions for model training, motivated by the cost, scarcity, and privacy limitations of human-collected data, as well as its susceptibility to bias and annotation noise \cite{honovich2023unnatural}. LLM-based data generation has been successfully applied across domains such as code generation \cite{luo2023wizardcoder} and instruction following \cite{wang2023self}, and has played a central role in training mainstream models including Alpaca \cite{taori2023stanford}. TeachingCoach builds on this line of work by using LLMs to synthesize multi-turn instructional dialogues and training on curated synthetic data filtered by experts.

Beyond their role as data generators, recent work has also explored the use of LLMs as interactive tutors and instructional agents for learning support. 

Recent work has examined LLMs for learning and tutoring, highlighting both their promise and limitations. Evaluations such as EducationQ show that teaching effectiveness does not scale linearly with model size or general reasoning ability \cite{shi2025educationq}. Systems such as PACE and Agent4Edu incorporate pedagogical strategies through Socratic questioning and learner modeling to enable personalized support \cite{liu2025pace,gao2025agent4edu}. However, these systems primarily focus on student-centered tutoring. In contrast, recent human-centered design research explores AI-based pedagogical agents that directly support instructors by fostering trust, social transparency, and flexible engagement, particularly to encourage adoption among AI-conservative educators \cite{chen2025bridging}. 

Related work has also examined classroom uses of generative AI with an emphasis on AI literacy. Studies with middle school English Language Arts teachers show how scaffolded AI use can support students' understanding of AI concepts and ethical use \cite{ritchie2025ela}, but position instructors mainly as facilitators rather than recipients of instructional support. Similarly, recent approaches that train LLM-based tutors using student models and pedagogical rubrics prioritize student outcomes and offer limited support for instructors' reflective practices or instructional decision-making \cite{scarlatos2025utterance}.

\section{Pipeline of TeachingCoach}\label{sec2}

\subsection{Rules Extraction}
We start by building a pedagogical knowledge base from authoritative education resources \cite{lang2021small,darby2019small}, where experts extract 36 core instructional rules and encode them as structured system prompts. Each rule functions as a guiding pedagogical principle in dialogue generation, ensuring that conversations remain anchored in evidence-based practices rather than drifting toward generic or unsupported advice.

\subsection{Data Collection}
To support realistic and diverse training data, we use GPT-4o \cite{hurst2024gpt} to generate \textbf{teacher profiles} that specify years of experience and teaching subject, as well as \textbf{teaching challenges} such as managing classroom attention. 
Given the extracted rules, teacher profiles, and challenges, GPT-4o produces \textbf{multi-turn conversations} between a teacher and a simulated expert. These dialogues capture how instructional principles may be applied in practice. To guarantee quality, human experts then review the conversations, removing those that are inconsistent, repetitive, or pedagogically unsound. The dataset serves as high-quality input for fine-tuning the model. After filtering, the dataset contains 406{,}183 training, 4{,}156 validation, and 4{,}143 test examples.

Specifically, each multi-turn conversation is generated by starting from a single validated teaching dilemma that implicitly violates one of the 36 practices. This scenario is paired with a synthetic instructor profile (e.g., course type, class size, teaching experience, and personality cues) to ground the interaction in a realistic classroom context. The conversation then unfolds as a dialogue between the instructor and an expert teaching consultant over 20–30 turns, progressing through stages of clarifying the teaching challenge, exploring possible instructional strategies, planning concrete next steps, and reflecting on how those strategies might work in practice. Instructor turns are conditioned on the scenario and profile to surface realistic concerns and follow-up questions, while expert responses are constrained to be concise, supportive, and actionable, mirroring authentic teaching-center consultations.

\subsection{Pedagogically-Grounded Model Training}

We fine-tune a LLaMA-2-13B-Chat model \cite{touvron2023llama} with full parameter update  on curated multi-turn dialogues. Each dialogue consists of a \textbf{system message} (teaching rules for the current instructional phase), a \textbf{user message} (profile and challenge), and an \textbf{assistant message} that first predicts a step descriptor and then generates the response. Steps follow the Eberly Center strategies\footnote{\url{ https://www.cmu.edu/teaching/solveproblem/index.html.}}--\textit{Step 1: Identify the Problem}, \textit{Step 2: Explore Reasons}, and \textit{Step 3: Develop Strategies}. Here is an example:
\begin{verbatim}
System: [rules keyed by phase]
User:   [profile + challenge]
Assistant: <step>Step 1</step> [response]
...
User:   [follow-up]
Assistant: <step>Step 2</step> [response]
\end{verbatim}

Let $H^{(i)}_t$ denote the full conversation history up to assistant turn $t$ in dialogue $i$ (including all prior system/user/assistant turns). The assistant is trained to (1) \emph{identify the current instructional step} $\sigma^{(i)}_t$ from $H^{(i)}_t$, and then (2) \emph{generate the response} $y^{(i)}_t$ conditioned on both $H^{(i)}_t$ and $\sigma^{(i)}_t$. We optimize the next-token objective over the entire assistant turn (step tokens + response tokens):

\[
\mathcal{L}(\theta) 
= - \sum_{i,t} \log P_\theta\!\big([\sigma^{(i)}_t, y^{(i)}_t] \mid H^{(i)}_t\big),
\]

At inference, the model first generates \texttt{<step>Step $m$</step>} by leveraging $H_t$ (the dialogue so far), thereby \emph{judging} which instructional stage applies; it then produces a response guided by the corresponding step-specific rules provided in the system prompt. For the end-user interface, the step tag can be logged for analytics and personalization while optionally being hidden from the visible transcript. 

Because the step label itself is generated by the model, TeachingCoach aligns dialogue progression with a structured pedagogical scaffold, producing conversations that naturally advance from problem identification to diagnosis to strategy development.

We chose LLaMA-13B as the backbone model because it provides sufficient capability to model structured, multi-turn instructional dialogue while remaining feasible to fine-tune on academic compute resources. Its open-weight nature enables full supervision over both instructional step predictions and response generation, which is essential for implementing pedagogically grounded training objectives. 

We compare against GPT-4o as a baseline because it represents a strong, widely deployed general-purpose LLM. Evaluating GPT-4o in a zero-shot setting highlights the impact of explicit pedagogical supervision and step-aware training in TeachingCoach, independent of model scale or proprietary data. 

\section{Evaluation}
We asked teaching experts to evaluate 200 conversations generated by \textit{TeachingCoach} and a GPT-4o baseline. Each dialogue turn was rated on four dimensions: \emph{E1 clarity of responses}, \emph{E2 respectful tone}, \emph{E3 encouragement of reflection and reasoning}, and \emph{E4 acknowledgment of user input}. Ratings used a 3-point scale (1 = poor, 3 = strong). Table~\ref{tab:evaluation} summarizes the results, showing that TeachingCoach consistently outperformed GPT-4o.

\begin{table}[t]
\centering
\resizebox{\columnwidth}{!}{%
\begin{tabular}{lcccc}
\hline
 & \textbf{E1} & \textbf{E2} & \textbf{E3} & \textbf{E4} \\
\hline
GPT-4o baseline & 2.01 & 1.99 & 1.68 & 1.85 \\
TeachingCoach & \textbf{2.62} & \textbf{2.55} & \textbf{2.59} & \textbf{2.56} \\
\hline
\end{tabular}
}
\caption{Average expert ratings over 200 conversations (3-point scale). Higher scores are marked in bold.}
\label{tab:evaluation}
\end{table}


\begin{figure}[ht!]
    \centering
    \includegraphics[width=0.95\linewidth]{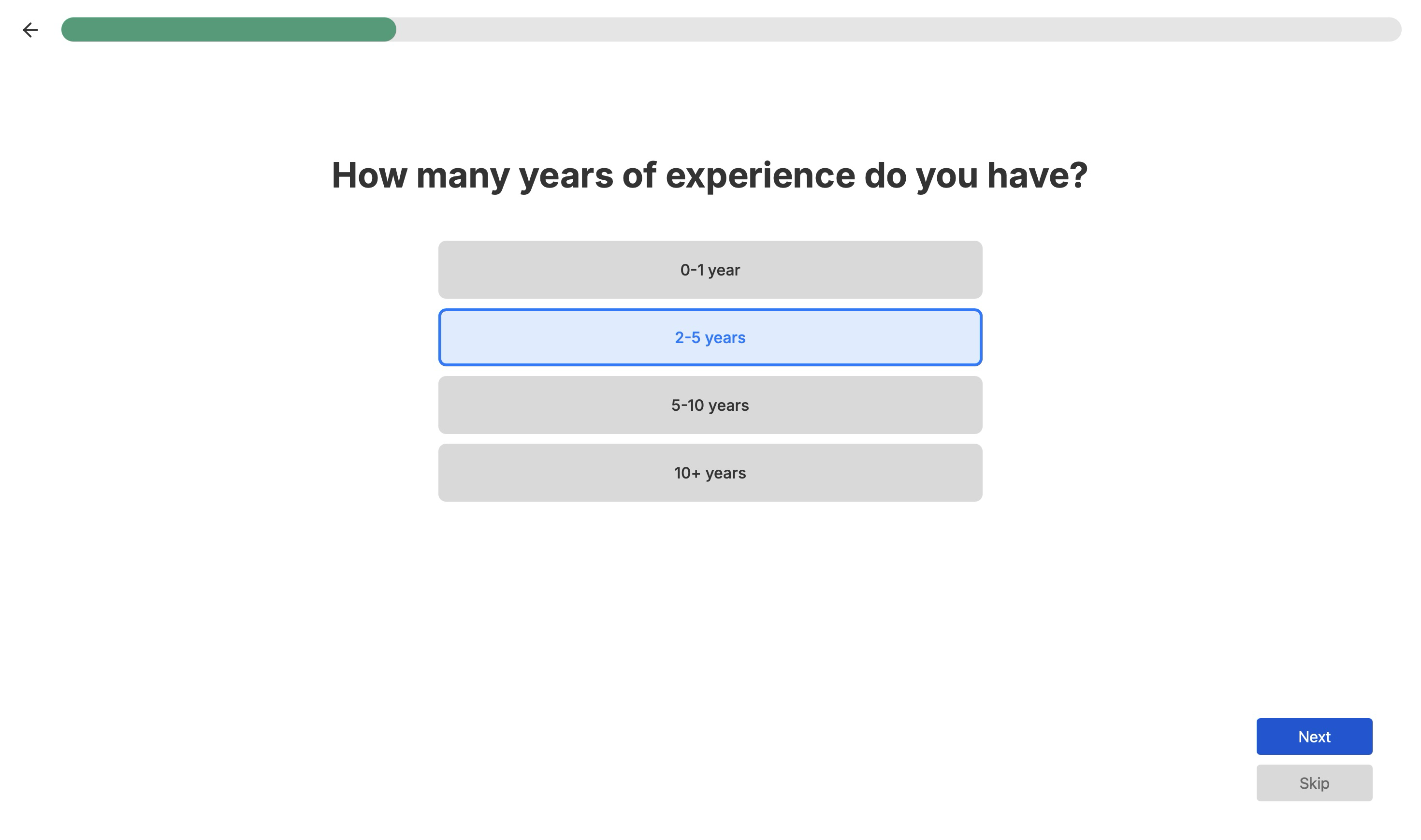}
    \caption{The onboarding asks users to specific their experience, current courses, and AI attitudes}
    \label{fig:your_label}
\end{figure}

\begin{figure}[t]
\centering
\includegraphics[width=0.95\columnwidth]{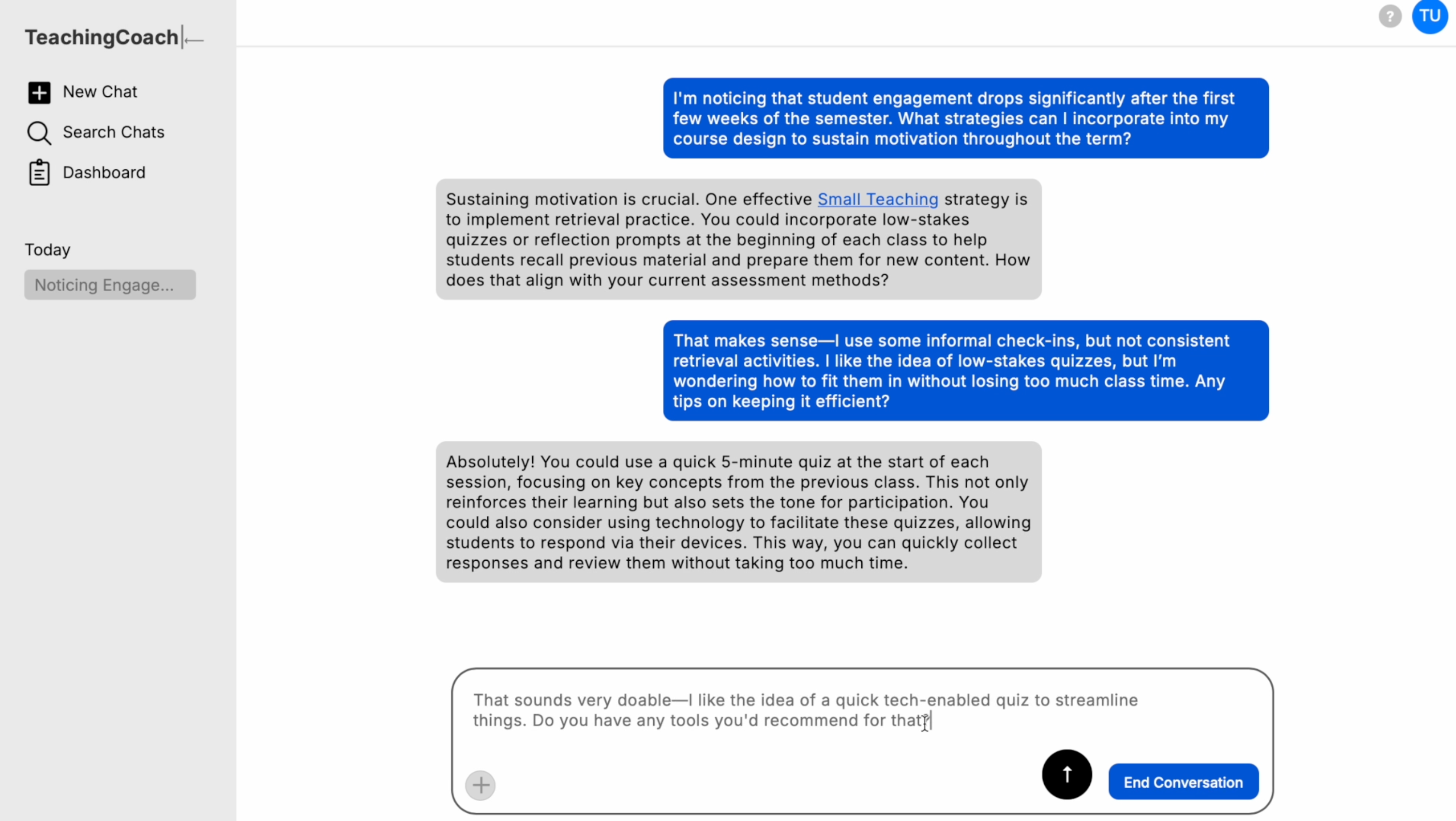}
\caption{The chatbot interface allows users to interact with the TeachingCoach model, demonstrating multi-turn instructional support.}
\label{fig:demo-ui}
\end{figure}

\begin{figure}[ht!]
    \centering
    \includegraphics[width=0.95\linewidth]{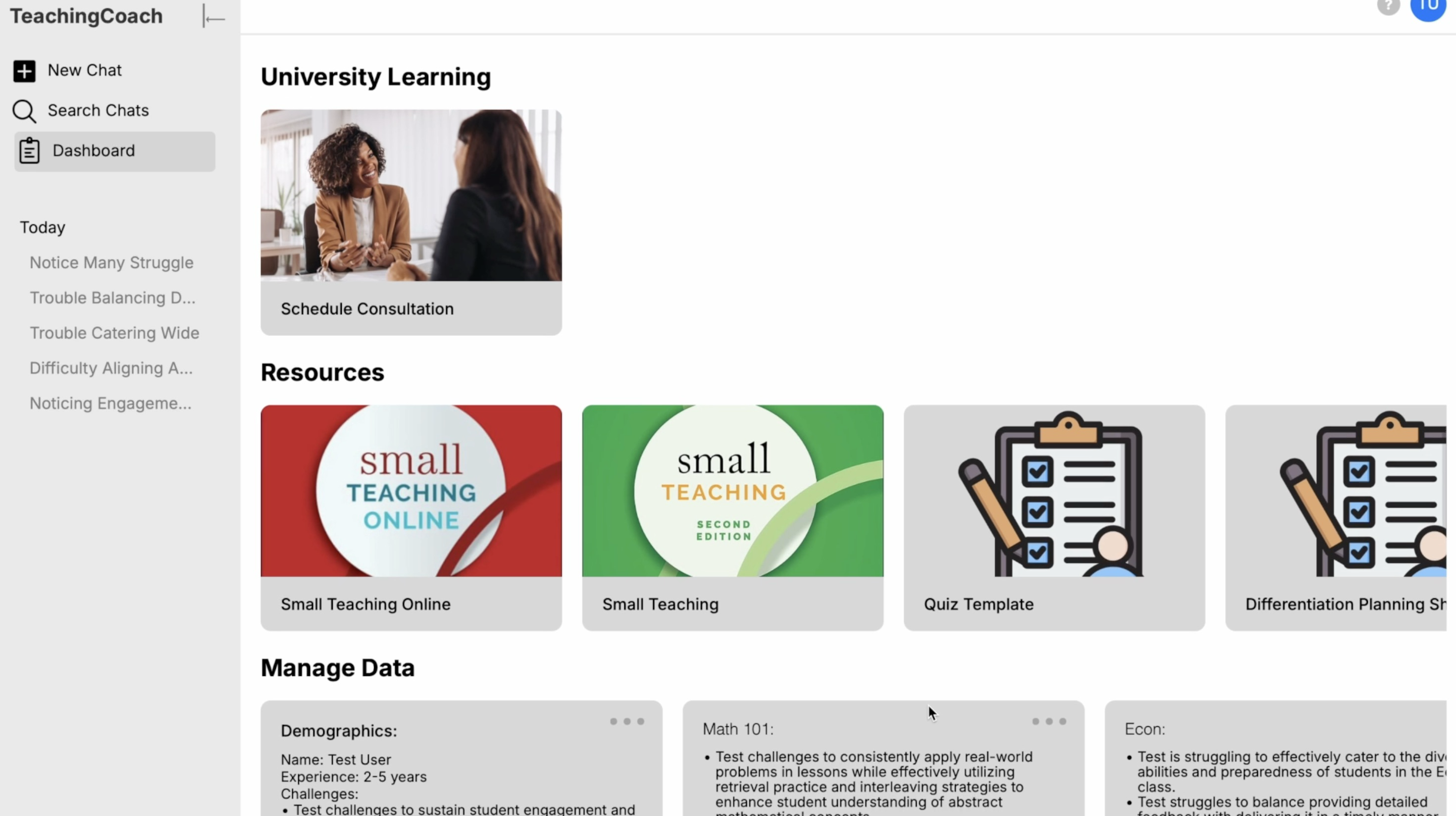}
    \caption{The dashboard provides a window for users to schedule consultations with live experts, view the resources they have collected, and manage their stored data}
    \label{fig:your_label}
\end{figure}

\section{Demo System}
TeachingCoach provides three components: \textbf{onboarding}, the \textbf{chatbot interface}, and the \textbf{dashboard}. Onboarding collects lightweight background information (e.g., experience, courses taught, key challenges), to enable personalization with minimal user burden. The chatbot interface is the core feature, supporting multi-turn conversations about teaching challenges. Dialogues follow a structured flow—problem identification, cause exploration, and strategy development—enabling targeted, pedagogically grounded suggestions. As shown in Figure~\ref{fig:demo-ui}, an instructor might raise a concern such as student engagement, and the chatbot responds with strategies like low-stakes quizzes or technology-enabled check-ins. The dashboard acts as a hub for reflection and resource management, storing conversation summaries, generated suggestions, and user data. Together, these components make TeachingCoach accessible, personalized, and scalable for professional support.

\section{Preliminary User Study and Results}
\subsection{Study Design}
We conducted a remote user study in August 2025 in which participants interacted with two versions of a demo system: one powered by \textit{TeachingCoach} and one by a GPT-4o-mini baseline. Participants explored the same teaching challenge with each system for up to 10 minutes per condition, with order counterbalanced. After each interaction, participants completed a brief survey and a 15-minute post-study interview. 

We recruited 41 higher education instructors through public postings across U.S. research universities, community colleges, and liberal arts colleges. The sample included 22 faculty members and 19 teaching assistants or instructional staff. Participants were informed that two versions of the system existed but were not told how the systems differed or which model powered each version. The session concluded with a semi-structured interview focused on perceived differences between systems, preferences, and perceived learning support.

\subsection{Findings}

The results show a distinction between overall preference and perceived recall. While the baseline model was preferred overall (21 vs. 13), participants more often attributed learning to the fine-tuned model (18 vs. 13). Agreement between preference and learning attribution was low (39\%), indicating that features driving preference do not always match perceived instructional value. Participants consistently distinguished the models on conversational engagement, efficiency of interaction, and breadth of suggestions. 

\textit{Conversational Engagement}.
The fine-tuned model was often described as engaging in dialogue resembling interaction with a human pedagogy expert. Participants noted its use of reflective questions and follow-up prompts that encouraged examination of teaching practices. This conversational depth supported learning through reflection, though some participants found it slowed the interaction when seeking quick answers.

\textit{Efficiency of Interaction}.
Efficiency was a key strength of the baseline model. Participants valued its ability to generate responses quickly, enabling rapid access to ideas when seeking immediate guidance. In contrast, the fine-tuned model often required more conversational turns to reach concrete suggestions, which some perceived as inefficient, though others found the slower pace helpful for clarifying goals and context.

\textit{Breadth of Suggestions}.
The baseline model was associated with greater breadth, offering wide-ranging strategies across contexts, which allowed users to scan for relevant ideas but several participants felt overwhelmed. By contrast, the fine-tuned model offered a narrower, more focused set of suggestions centered on specific classroom practices, improving coherence while reducing exposure to diverse strategies. 

Overall the findings highlight a trade-off between depth and efficiency: the fine-tuned model supported learning through sustained engagement and reflection, while the baseline model prioritized speed and breadth of suggestions.  

\section{Discussion and Conclusion}
\textit{TeachingCoach} demonstrates how LLMs can be trained to follow explicit instructional guidance by encoding instructional steps directly into training dialogues. Rather than relying on large-scale human-collected teaching conversations, we use fully synthetic data generated from instructional guidelines, instructor profiles, and teaching challenges, enabling scalable training while avoiding privacy and ethical concerns. 

To understand how this step-aware instructional behavior is perceived in practice, we conducted a user study comparing TeachingCoach with a GPT-4o-mini baseline, which represents a
widely used general-purpose conversational model optimized for efficiency and broad coverage. The study reveals a clear trade-off between interaction efficiency and instructional depth: while the baseline model was often valued for its speed and breadth of suggestions, TeachingCoach was more frequently associated with deeper instructional support through reflective dialogue and structured reasoning. 

Rather than viewing this trade-off as a limitation, it motivates the development of hybrid instructional systems that balance breadth and depth based on user needs and context. One promising direction is a multi-agent architecture, in which agents are specialized for complementary roles—for example, one prioritizing rapid idea generation and broad strategy exploration, and another providing structured instructional guidance and reflective questioning. Hybrid behavior can also be achieved within a single agent through adaptive interaction modes, such as beginning with concise, breadth-oriented suggestions and transitioning into deeper instructional guidance when users seek reflection or clarification. In addition, explicit control options (e.g., quick suggestions versus guided reflection) or progressive disclosure strategies can allow instructors to align system behavior with their immediate instructional goals.

This study has several limitations. The sample comprised self-selected higher education instructors, which may limit generalizability to K-12 settings, informal learning environments, or institutions with different cultural and technological infrastructures. Additionally, the evaluation focused on instructors' perceptions and did not directly assess student learning outcomes; future work should examine the effects of step-aware instructional dialogue on classroom practice and learning over time. 

Although our study focuses on higher education instructors, the data generation pipeline and training formulation are not limited to this context. By modifying instructor profiles, teaching challenges, and contextual constraints, the same approach can be applied to other instructor-facing settings, such as K–12 teaching, professional training, instructional coaching, mentoring, and tutoring support. More broadly, this work points toward a general approach for developing instructor-facing instructional dialogue systems that support reflective teaching practice across domains, while preserving pedagogical consistency, scalability, and privacy. 

Please find conversation examples and our codebase at \url{https://osf.io/n2xyu/overview?view_only=e5b85d85b4a842dea902d9714f6faa67}.

\bibliographystyle{ACM-Reference-Format}
\bibliography{references}


\end{document}